\documentclass[conference]{IEEEtran}

\usepackage[pdftex]{graphicx}
\usepackage{amssymb}
\graphicspath{{./figures/}}
\usepackage[cmex10]{amsmath}
\usepackage{url}
\usepackage{stfloats}
\usepackage{graphicx}
\usepackage{multirow}
\usepackage{array}

\begin{document}
\title{Food Image Classification and Segmentation \\ with Attention-based Multiple Instance Learning}

\author{\IEEEauthorblockN{Valasia Vlachopoulou,
Ioannis Sarafis, and
Alexandros Papadopoulos}
\IEEEauthorblockA{Department of Electrical and Computer Engineering, \\
Aristotle University of Thessaloniki, Greece \\
Email: vlachopvs@ece.auth.gr, sarafis@mug.ee.auth.gr, alpapado@mug.ee.auth.gr}
}

\maketitle

\IEEEpubidadjcol

\begin{abstract}
The demand for accurate food quantification has increased in the recent years, driven by the needs of applications in dietary monitoring. At the same time, computer vision approaches have exhibited great potential in automating tasks within the food domain. Traditionally, the development of machine learning models for these problems relies on training data sets with pixel-level class annotations. However, this approach introduces challenges arising from data collection and ground truth generation that quickly become costly and error-prone since they must be performed in multiple settings and for thousands of classes. To overcome these challenges, the paper presents a weakly supervised methodology for training food image classification and semantic segmentation models without relying on pixel-level annotations. The proposed methodology is based on a multiple instance learning approach in combination with an attention-based mechanism. At test time, the models are used for classification and, concurrently, the attention mechanism generates semantic heat maps which are used for food class segmentation. In the paper, we conduct experiments on two meta-classes within the FoodSeg103 data set to verify the feasibility of the proposed approach and we explore the functioning properties of the attention mechanism.  
\end{abstract}

\section{Introduction} 
\label{sec:introduction}
As the world’s population percentage that suffers from the prevalence of overweight and obesity increases, the role of digital self-monitoring of diet, as well as of physical activity, has a significant effect in maintaining a healthy weight, with tailored, personalised advising having a key role in effective interventions \cite{berry_self-monitoring_2021}. Computer vision methods for dietary motoring have seen increased use for the tasks of food recognition and volume estimation, with a rapidly growing number of technology-assisted dietary studies relying on deep learning models to perform food recognition \cite{tahir_a-comprehensive-survey_2021}. Nevertheless, challenges persist in this domain, including the ones caused by occlusions, illumination, shape, and structure variation of food \cite{chopra2022recent} as well as differences from food preparation and cooking styles across regions or cultures \cite{tahir_a-comprehensive-survey_2021}. While the demand for large-scale and diverse food image data sets is growing, and large-scale benchmarks are becoming available, the availability of pixel-level annotations remains fairly scarce \cite{min_large-scale_2023}. 

In relation to the problem of dietary monitoring, Yogaswara et at. \cite{yogaswara_instance-aware_2019} proposed the use of instance-aware semantic classification and segmentation for the task of estimating the caloric content of images captured by a smartphone camera. In their approach, they first use the Mask Region-based Convolutional Neural Networks (Mask R-CNN) \cite{He_2017_ICCV} to identify different instances of each object, then they estimate the area and the volume of each food type, and finally they provide an estimation of the caloric content. The models are trained  and evaluated on images with pixel annotations taken in a controlled lab environment, where the different food types/instances do not overlap in the plates. Freitas et al. \cite{freitas_myfood_2020} proposed a nutritional monitoring system based on classification and segmentation accessible via a smartphone app. In their experiments, they trained five supervised learning models using pixel annotations and evaluated their performance on Brazilian food types, with Fully Convolutive Networks (FCN) \cite{He_2017_ICCV} and Mask R-CNN achieving the top performance in the target domain. Similarly, Wu et al.\cite{10.1145/3474085.3475201} evaluate their method called ReLeM against common baselines for the task of semantic segmentation, including Dilated Convolution based \cite{Huang_2019_ICCV}, Feature Pyramid based \cite{Kirillov_2019_CVPR}, and Vision Transformer based \cite{Zheng_2021_CVPR}, all of which require pixel-level annotations for training images.

Contrary to previous approaches, \emph{Weakly Supervised Machine Learning (WSML)} methodologies reduce reliance on pixel-level annotations and have been successfully employed for image segmentation in other domains, such as medical imaging \cite{papadopoulos2021interpretable, Chen_2022_CVPR, yu_adaptive-soft_20022, xu_camel_2019}. In the food domain, the seminal work of Shimoda and Yanai \cite{shimoda_cnn-based_2015} uses CNNs trained on image-level annotations – without use of pixel-level annotations – and the predictions are combined with the GrabCut algorithm \cite{rother_grabcut_2004} to identify regions for various food categories. Their objective is not to provide precise segmentations, but to achieve a good bounding box with at least $50\%$ overlap with the ground truth. Notably, subsequent approaches of WSML in the food domain have focused in identifying the area of food (i.e., to distinguish it from the plate and other background), rather than  classification and segmentation for separate food classes \cite{wang_weakly_2017, shimoda_weakly_2020}. 

To overcome the restrictions of the previous approaches, in this paper we present and evaluate an \emph{attention-based}, \emph{multiple instance learning (MIL)} methodology for semantic food image classification and segmentation, applicable to distinct food classes. In each model, the attention mechanism is used to detect the target food class and to create a heatmap that highlights the areas of the class in the images. Image segmentations are then provided by applying a threshold on the heatmap values. Experiments conducted on two meta-classes created within the FoodSeg103 \cite{10.1145/3474085.3475201} data set demonstrate the feasibility of our approach.

The rest of the paper is organised as follows. Section \ref{sec:methodology} first presents the methodology for model training, then explains how to acquire class predictions and semantic segmentations at test time. Section \ref{sec:dataset} gives an overview of the used data set and describes the meta-classes created from it. Then, Section \ref{sec:experiments} describes the experimental details, presents the classification results and provides illustrations of the heatmaps and segmentation produced by the attention-mechanism at test time. Finally, Section \ref{sec:conclusions} concludes the paper.

\section{Proposed Methodology} 
\label{sec:methodology}
\begin{figure*}[!t]
\centering
 \includegraphics[width=6in]{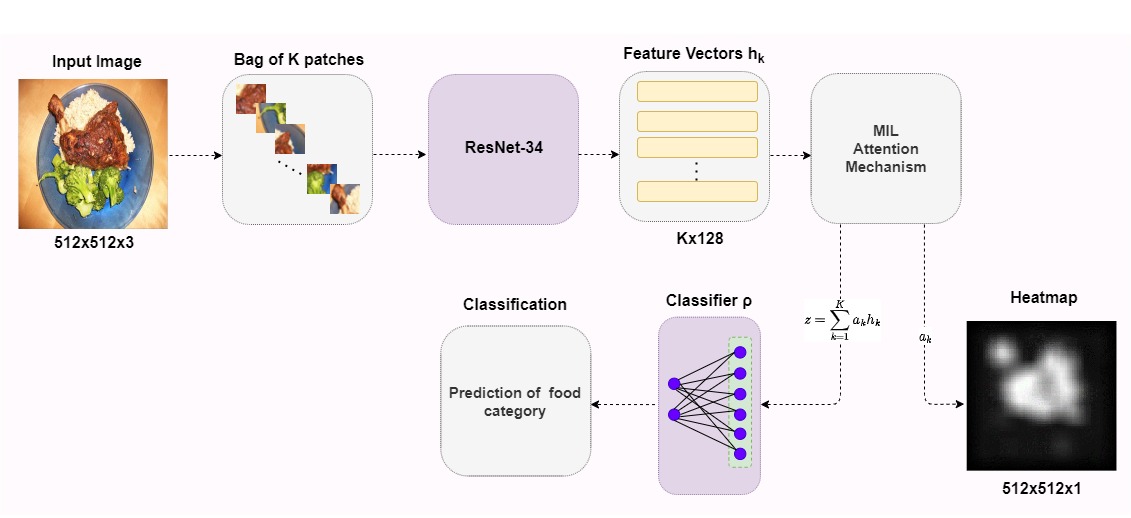}
\caption{\label{figure:overview-methodology} High-level overview of the proposed methodology, previously used for diabetic retinopathy detection \cite{papadopoulos2021interpretable}. A multiple instance model with attention is trained for the classification problem for the target food class, without the use of pixel-level annotations. At test time, the model predictions are used to detect whether the target food class is available in the image and, concurrently, the attention mechanism is used to produce heatmaps that localise the area of the target class in the test images.}
\end{figure*}

Figure \ref{figure:overview-methodology} illustrates on overview of the model's architecture and methodology which is used to train binary classification models. At test time, the models  detect the existence of a target food class and, simultaneously, perform semantic segmentation for this class using the heatmaps generated by accumulating the output of the attention mechanism. 

\subsection{Model Architecture and Model Training}

The proposed methodology is based on the work of Papadopoulos et al. \cite{papadopoulos2021interpretable}, who tackled the detection of referable diabetic retinopathy in eye fundus images by using the paradigm of multiple-instance learning. In their work, they demonstrated that their approach achieved high classification performance. Additionally, they explored the idea of providing interoperability of the results based on the heatmaps produced using the attention mechanism. 

Briefly, the network's architecture consists of three components: a Residual Network 34 (ResNet-34) \cite{7780459}, an attention mechanism \cite{pmlr-v80-ilse18a} and a classifier responsible for categorizing the original images into distinct food categories. Within the framework of multiple instance learning, every training instance is represented as a \emph{bag of patches} extracted from the original image, and the labels are defined in the bag level. A bag is labeled as positive, if its constituent patches originate from an image that depicts food of the target class. Conversely, it is labeled as negative when its patches do not contain any food of the target category. 

Specifically, each image is divided using square patches of dimensions $d\times d$ with an overlap rate $t \in [0,1)$. During training, for each training image, $K$ random patches $\{x_{1}, x_{2},...,x_{K}\}$, $x_{k} \in \mathbb{R}^{d\times d\times 3}$, are selected from the set of all possible patches to form the bag $X$, that will be fed into the network. Since each image is represented using only a small subset of the available patches, this technique enhances the training acceleration and the data variation.  Additionally, between epoch iterations the bag for an image will contain different random patches. Thus, this approach augments the training set, making the model training more robust to potential overfitting challenges.   

After the bag creation from the original images, each bag  passes through the ResNet-34 network and we use its last layer for feature extraction. This leads to the transformed set of vectors $H = \{h_{1}, h_{2},...,h_{K}\}$, containing $K$ feature vectors of $M = 128$ dimensions, which serve as input to the attention mechanism. 

The attention mechanism used in this network was inspired by Ilse et al. \cite{pmlr-v80-ilse18a}, who proposed a trainable weighted mean to obtain a one-dimensional, image embedding vector $z \in \mathbb{R}^{M}$. The attention weights are determined by a neural network consisting of two fully connected layers with the trainable parameters represented by the vectors $w$ and $V$. Furthermore, the attention weights are normalized into the range $[0, 1]$ and sum up to one, ensuring that the mechanism remains invariant to changes in bag dimensions. Specifically, $z$ is calculated as:
\[z = \sum_{k=1}^{K} a_{k}h_{k} \tag{1} \label{eq1}\] where
\[a_{k} = \frac{\text{exp}(w^T \text{tanh}(Vh_{k}^T))}{\sum_{j=1}^{K} \text{exp}(w^T \text{tanh}(Vh_{j}^T))} \tag{2} \label{eq2}\]
and
$w \in \mathbb{R}^{L\times 1}$, $V \in \mathbb{R}^{L\times M}$.

Subsequently, the embedding $z$, obtained after the attention mechanism, is propagated to the classifier $\rho: \mathbb{R}^{M} \rightarrow \mathbb{R}$, which is a single fully connected layer. The output is a scalar which, after being propagate via a sigmoid function, represents the probability that the original image belongs to the positive food category. 

In a nutshell, the network determines this prediction probability as:
\[p(y|X) = \sigma\left(\rho(z)\right) \tag{3} \label{eq:predict-proba}
\]
where 
\[
 \rho(z) = \rho\left(\sum_{k=1}^{K} a_{k}h_{k}\right) = \rho\left(\sum_{k=1}^{K} a_{k}\phi(x_{k})\right) \tag{4}
 \]
with $\sigma$ denoting the sigmoid function, $\phi$ denoting the feature extraction of ResNet-34, and the weights $\{a_k\}$ calculated by the attention mechanism.

\subsection{Classification and Segmentation for New Images}
Regarding the classification of a new, test image, the network generates the prediction $\hat{y}$ at the image level. Considering the binary nature of the problem, the classifier's output for a test image is the probability $y_{prob} = \mathbb{P}(y=1|X)$ that the image belongs to the positive class. The final step involves converting this probability into a predicted label, $\hat{y}$, by comparing it to the threshold of $0.5$ .

For the segmentation of a test image, the bag of patches is generated using a different policy than when in training. Particularly, each bag is formed by densely sampling square patches throughout the image. Again, square patches with the same dimensions, $d\times d$, are used but with an a higher overlap rate of $t' > t $. Consequently, the count $K_{test}$ of patches in an image with dimensions $D \times D$ is determined as:
\[K_{test} = \left\lceil \left(1 +\frac{D-d}{d(1-t')} \right)^2 \tag{5} \right\rceil \]

Then, for each test image, we create a heatmap that localises the predicted class by accumulating the attention scores for each patch. The heatmap has the same dimensions as the input image, $D \times D$, and is constructed based on the attention weight values, $a_{k}$, assigned to each part of the image within the densely-sampled bag. More precisely, each weight $a_{k}$ is normalized and aggregated in the pixel coordinates of the heatmap from which each patch originates. Finally, a threshold $a$ is applied on the heatmap values to determine the segmentations.

Notably, as the overlap rate $t'$, the number of patches $K_{test}$ also increases and the heatmap tends to be more informative at the pixel level, thus leading to a more granular and comprehensive final representation and resulting segmentations.

\section{Data set}
\label{sec:dataset}
For our experiments, we use the FoodSeg103 data set \cite{10.1145/3474085.3475201} which encompasses 103 distinct food categories in a set of 7,118 images with pixel-level annotated segmentation masks. Each image contains on average six different food categories. The authors divided the images into training and test sets, maintaining a ratio of 7:3 between them. Overall, the training set consists of 4,983 images, accompanied by 29,530 segmentation masks, while the test set includes 2,135 images with 12,567 segmentation masks. The same train and test sets are used in our experiments. It is worth noting, however, that in this work the segmentation masks in the training set are solely used to create binary classes at image level.

Given that the data set contains numerous food categories, we have the limitation that there are not abundant images of each individual class in the training set. Moreover, considering the MIL aspect, the difficulty of the problem increases. Consequently, we formed meta-classes in FoodSeg103, by merging relevant individual classes into broader ones. The two meta-classes created are referred to as "Bakery" and "Meat". The "Bakery" contains the individual categories: egg tart, biscuit, cake, bread, while the "Meat" includes the classes: steak, pork, chicken duck, sausage, fried meat, lamb. After the consolidation, the total number of images for "Bakery" turned out to be 2.163 and for "Meat" 3.562 images. The distribution in the individual categories are presented in Tables \ref{table:bakery-class} and \ref{table:meat-class}.

\begin{table}[!h]
\renewcommand{\arraystretch}{1.3}
\caption{Class distribution in the "Bakery" meta-class.}
\label{table:bakery-class}
\centering
\begin{tabular}{c|c|c|c}
\hline
\bfseries Bread & \bfseries Cake & \bfseries Biscuit & \bfseries Egg tart\\
\hline\hline
1.405 & 459 & 295 & 4 \\
\hline
\end{tabular}
\end{table}

\begin{table}[!h]
\renewcommand{\arraystretch}{1.3}
\caption{Class distribution in the "Meat" meta-class.}
\label{table:meat-class}
\centering
\begin{tabular}{c|c|c|c|c|c}
\hline
\bfseries Steak & \bfseries Pork & \bfseries Chicken duck & \bfseries Sausage & \bfseries Fried meat & \bfseries Lamb\\
\hline\hline
1.065 & 669 & 1.242 & 249 & 238 & 99 \\
\hline
\end{tabular}
\end{table}

A common pre-processing stage is applied to all images before they are entered into the network during training or testing. Since the size of the patches forming the bags remains constant throughout each training iteration, it is essential to maintain uniformity in the image dimensions. Therefore, all images are resized to dimensions $512 \times 512$ using bilinear interpolation. 

Having established a common reference point concerning the size, the next stage involves excluding certain images both from the training and test sets, based on a threshold for the number of pixels belonging to the positive class. Specifically, when the total number of pixels belonging to the positive category in an image is less than $20.000$, the image is discarded and not utilized by the network at any stage. That is, given the total number of pixels of each image after resizing to  $512 \times 512$, an image is considered "positive" if the target food class covers at least $7.6\%$ of the image area.

\section{Experiments}
\label{sec:experiments}
We conducted two experiments to evaluate the effectiveness of the proposed method for semantic classification and segmentation of the target meta-classes "Meat" and "Bakery" within food images. 

During the training process, for the "Meat" category we utilized 2,048 positive images and 2,627 negative images. Similarly, for the "Bakery" category, the  training set consisted of 1,330 positive instances and 3,464 negative instances. Given that the ratio of positive to negative class for the "Bakery" meta-class is approximately $1:3$, we employed oversampling of the positive images to achieve class-balanced mini-batches. For testing, the "Meat" test set includes 917 positive images and 2,015 negative images, while the test set for the category "Bakery" contains $543$ positive and 1,514 negative images.  

Regarding the training parameters, each input image is transformed to its bag of patches representation by randomly selecting $K = 50$ random patches with dimensions $64 \times 64$ from the pool of image patches created with an overlap $t=75\%$. The model is trained for 130 epochs in total. We initialized ResNet-34 with pre-trained ImageNet weights. The ResNet-34 weights are immutable for the initial 50 epochs and are fine-tuned in the subsequent 80 epochs. In contrast to train time, the input image representation in testing is resulting from a patch extraction with same dimensions $64 \times 64$ but on a regular grid with higher overlap rate of $0.875$. 

\begin{figure}[!b]
\includegraphics[width=3.5in]{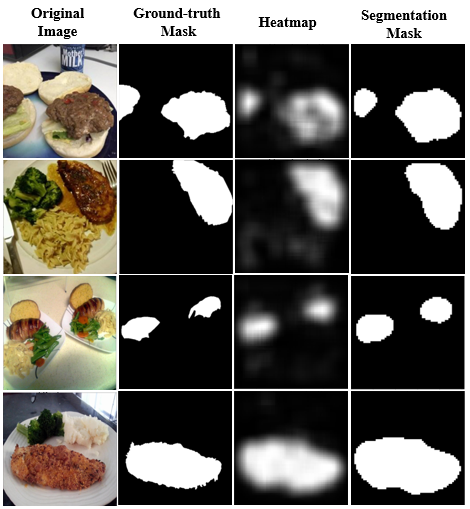}
\caption{Indicative examples of heatmaps and segmentation masks for positive images in the "Meat" meta-class. The first and second column contain the resized RGB images and the corresponding resized binary ground-truth masks, respectively. The third column displays the generated heatmaps, while the fourth column shows the segmentation masks, after having applied the threshold $a=0.3$ to the heatmaps' pixel values.}
\label{figure:visualizations_meat}
\end{figure}

First we evaluate the models for the task of image-level classification. Table \ref{tab:confusion_matrix_meat} presents the confusion matrix for the "Meat" class and Table \ref{tab:confusion_matrix_bakery} summarizes the confusion matrix for the "Bakery" category. The quantitative results of accuracy, precision, recall and F1-score for both meta-classes, are reported in Table \ref{tab:metrics_both}.

\begin{table}[!ht]
    \centering
    \renewcommand{\arraystretch}{1.3}
    \caption{The confusion matrix for class "Meat"}
    \label{tab:confusion_matrix_meat}
    \begin{tabular}{>{\centering\arraybackslash}m{1cm}>{\centering\arraybackslash}m{1cm}>{\centering\arraybackslash}m{1cm}>{\centering\arraybackslash}m{1cm}}
        \cline{2-4}
        \multicolumn{2}{c}{} & \bfseries Positive  & \bfseries Negative \\ \cline{2-4}
        
        \multirow{2}{*} & \bfseries Positive & 711 & 206 \\ \cline{2-4}
        & \bfseries Negative & 193  & 905 \\
        \cline{2-4}
    \end{tabular}
\end{table} 

\begin{table}[!ht]
    \centering
    \renewcommand{\arraystretch}{1.3}
    \caption{The confusion matrix for class "Bakery"}
    \label{tab:confusion_matrix_bakery}
    \begin{tabular}{>{\centering\arraybackslash}m{1cm}>{\centering\arraybackslash}m{1cm}>{\centering\arraybackslash}m{1cm}>{\centering\arraybackslash}m{1cm}}
        \cline{2-4}
        \multicolumn{2}{c}{} & \bfseries Positive  & \bfseries Negative \\ \cline{2-4}
        
        \multirow{2}{*} & \bfseries Positive & 287 & 256 \\ \cline{2-4}
        & \bfseries Negative & 58  & 1466 \\
        \cline{2-4}
    \end{tabular}
\end{table}

\begin{figure}[!t]
\includegraphics[width=3.5in]{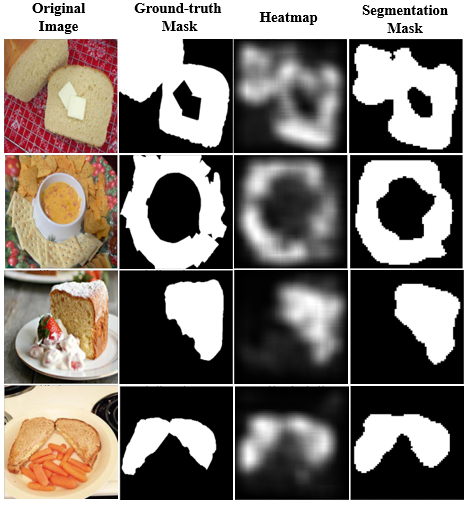}
\caption{Indicative examples of heatmaps and segmentation masks for positive images in the "Bakery" meta-class. The first and second column contain the resized RGB images and the corresponding resized binary ground-truth masks, respectively. The third column displays the generated heatmaps, while the fourth column shows the segmentation masks, after having applied the threshold $a=0.3$ to the heatmaps' pixel values.}
\label{figure:visualizations_bakery}
\end{figure}

In the second stage, the trained models are evaluated pixel-wise for the generated heatmaps, in terms of localizing the region of the target food classes in the positive images of the test sets. The metric of Average Precision (AP) evaluates the ranking of the pixels based on the values contained in the heatmaps, while the metric Intersection over Union (IoU) is computed by comparing the ground-truth masks of the images with the corresponding heatmaps, after applying the threshold of $a=0.3$ in the heatmaps' pixel values. Table \ref{tab:metrics_both} reports the results of IoU and AP for the categories "Meat" and "Bakery".

\begin{table}[!ht]
    \centering
    \renewcommand{\arraystretch}{1.3}
    \caption{Classification and segmentation metrics for "Meat" and "Bakery"}
    \label{tab:metrics_both}
    \begin{tabular}{>{\centering\arraybackslash}m{0.8cm}>{\centering\arraybackslash}m{0.8cm}>{\centering\arraybackslash}m{0.8cm}>{\centering\arraybackslash}m{0.8cm}>{\centering\arraybackslash}m{0.8cm}|>{\centering\arraybackslash}m{0.8cm}>{\centering\arraybackslash}m{0.8cm}}
        \cline{1-7}
        \multicolumn{1}{c}{} & \bfseries Accuracy  & \bfseries Precision & \bfseries Recall & \multicolumn{1}{c|}{\bfseries F1-score}  & \bfseries IoU & \bfseries AP \\ \cline{1-7}
        
        \multicolumn{1}{c}{\bfseries Meat} & 80.2\% & 78.6\% & 77.5\% & \multicolumn{1}{c|}{78\%} & 53.4\% & 77.5\% \\ \cline{1-7}
        \multicolumn{1}{c}{\bfseries Bakery} & 84.8\%  & 83.1\% & 52.8\% & \multicolumn{1}{c|}{64.5\%} & 47.4\% & 71.4\% \\ \cline{1-7}
    \end{tabular}
\end{table}

Figure \ref{figure:visualizations_meat} presents illustrative images from the test set depicting food of the class "Meat," which the final model accurately classified as positive instances. The testing images from FoodSeg103 (first column) and the respective ground-truth masks (second column) are shown after the resizing transformation of $512 \times 512$, along with the corresponding heatmaps (third column). The segmentation masks (fourth column) are displayed after the applied threshold $a=0.3$ to the heatmaps' pixel values. Similarly, Figure \ref{figure:visualizations_bakery} presents indicative images from the test set of the category "Bakery" that the trained model accurately predicted as positive instances. 

\section{Conclusions} 
\label{sec:conclusions}
This paper demonstrates a multiple-instance, attention-based methodology for food image classification and semantic segmentation. The weakly supervised models are trained using only class labels for the images and without any individual pixel annotations. For new, unseen images the trained models are able to achieve high classification performance and, notably, they are able to localise the area of the target food class using the heat map produced by the attention mechanism. The results show that the used model architecture, previously evaluated for diabetic retinopathy classification in eye fundus images, can be extended to perform semantic segmentation in the food domain. Moreover, the results suggest that weakly supervised machine learning approaches can address the complex tasks of semantic food image classification and segmentation, which are crucial for dietary monitoring and other applications in the food domain. In the future, we plan to extend our methodology and produce models for more specific food categories (e.g., the ones that are included in each meta-class) and other types of food images.  

\section*{Acknowledgment}
This work is part of a project that has received funding from the European 
Union’s Horizon 2020 research and innovation programme under grant agreement
No 965231.

\bibliographystyle{IEEEtran}
\bibliography{refs}

\end{document}